\pdfoutput=1

\documentclass[11pt]{article}

\usepackage{EMNLP2023}

\usepackage{times}
\usepackage{latexsym}
\usepackage{tabularx}

\usepackage[T1]{fontenc}

\usepackage[utf8]{inputenc}

\usepackage{microtype}
\usepackage{fancyhdr}
\usepackage{inconsolata}
\usepackage{pgfplots}

\usepackage{lipsum}
\newcommand\blfootnote[1]{%
  \begingroup
  \renewcommand\thefootnote{}\footnote{#1}%
  \addtocounter{footnote}{-1}%
  \endgroup
}

\title{nlpBDpatriots at BLP-2023 Task 2: A Transfer Learning Approach to Bangla Sentiment Analysis}

\author{Dhiman Goswami\textsuperscript{*}, Md Nishat Raihan\textsuperscript{*}, Sadiya Sayara Chowdhury Puspo\textsuperscript{*}, \\ \textbf{Marcos Zampieri} \\
        George Mason University \\
         \texttt{\{dgoswam, mraihan2, spuspo, mzampier\}@gmu.edu} \\
        }

\begin{document}
\maketitle
\begin{abstract}

In this paper, we discuss the nlpBDpatriots entry to the shared task on Sentiment Analysis of Bangla Social Media Posts organized at the first workshop on Bangla Language Processing (BLP) co-located with EMNLP. The main objective of this task is to identify the polarity of social media content using a Bangla dataset annotated with positive, neutral, and negative labels provided by the shared task organizers. Our best system for this task is a transfer learning approach with data augmentation which achieved a micro F1 score of 0.71. Our best system ranked $12^{th}$ among 30 teams that participated in the competition. 
\blfootnote{*These three authors contributed equally to this work.} 
\blfootnote{\bf WARNING: This paper contains examples that are offensive in nature.}
\end{abstract}

\section{Introduction}

NLP has become a major domain of modern computational research, offering a lot of applications from machine translation to chatbots. However, much of this research has been concentrated on English and other high-resource languages like French, German, and Spanish. 

Bangla, despite being the seventh most spoken language in the world with approximately 273 million speakers \cite{ethnologue2023}, has not received similar attention from the NLP community. This gulf is not just an academic oversight; it has real-world implications. Bangla is a language of significant cultural heritage and economic activity. The development of NLP technologies for Bangla is both a scientific necessity and a practical imperative. The limited availability of Bangla NLP resources has led to a reliance on traditional machine learning techniques like SVMs and Naive Bayes classifiers for classification tasks such as sentiment analysis. The advent of deep learning models has opened new avenues. Models like BERT \cite{devlin2019bert} have shown promising results in languages other than English and has been recently trained to support Bangla \cite{kowsher2022bangla}. 

Sentiment analysis is increasingly becoming a vital tool for understanding public opinion and people's behavior \cite{rosenthal2017semeval}. It has found applications in various sectors, including finance, where it helps investors to leverage social media data for better investment decisions \cite{mishev2020evaluation}. In the context of Bangla, the utility of sentiment analysis extends beyond mere academic interest. It can serve as a powerful tool for businesses to gauge customer satisfaction, for policymakers to understand public sentiment, and even for social scientists studying behavioral trends.

In this paper, we evaluate several models and implement transfer learning for the shared task on Sentiment Analysis of Bangla Social Media Posts organized at the first workshop on Bangla Language Processing (BLP-2023) \cite{blp2023-overview-task2}. Moreover, an ensemble model consisting of three transformer-based models generates a superior performance over the other approaches.

\section{Related Work}

\paragraph{Initiating Sentiment Analysis in Bangla} Sentiment analysis, which was mainly focused on English (e.g. \citeauthor{yadav2020sentiment} \citeyear{yadav2020sentiment}, \citeauthor{saberi2017sentiment} \citeyear{saberi2017sentiment}), is now becoming popular in other low resource languages like Urdu (e.g. \citeauthor{noor2019sentiment} \citeyear{noor2019sentiment}, \citeauthor{muhammad2023innovations} \citeyear{muhammad2023innovations}), Pashto (e.g. \citeauthor{iqbal2022sentiment} \citeyear{iqbal2022sentiment}, \citeauthor{kamal2016pashto}, \citeauthor{kamal2016pashto}), Bangla (e.g. \citeauthor{islam2020sentiment} \citeyear{islam2020sentiment}, \citeauthor{akter2021bengali} \citeyear{akter2021bengali}). Researchers are actively working to improve how people analyze and modify Bangla online comments using different methods and datasets. They are doing a variety of tasks, from classifying documents to mining opinions and analyzing sentiment, all while adapting their techniques to the specifics of the Bangla language. For example, for document classification, \citet{rahman2020bangla} presented an approach using the transformer-based models BERT and ELECTRA with transfer learning. The models were fine-tuned on three Bangla datasets. Similarly, \citet{rahman2020bangla} explored character-level deep learning models for Bangla text classification, testing Convolutional Neural Networks (CNN) and Long Short-Term Memory (LSTM) models. On the other hand, for opinion mining, \citet{haque2019opinion} analyzed Bangla and Phonetic Bangla restaurant reviews using machine learning on a dataset of 1500 reviews. SVM achieved the highest accuracy of 75.58\%, outperforming prior models. 

\paragraph{Advancements of Sentiment Analysis in Bangla}\citet{islam2020sentiment} presented two new Bangla sentiment analysis datasets which achieved state-of-the-art results with multi-lingual BERT (71\% accuracy for 2-class, 60\% for 3-class), and notes sentiment differences in newspaper comments. \citet{tuhin2019automated} proposed two Bangla sentiment analysis methods: Naive Bayes and a topical approach, aiming at six emotions, which achieved over 90\% accuracy for sentence-level emotion classification, outperforming Naive Bayes. Similarly, \citet{al2021social} discussed research focused on sentiment analysis and hate speech detection in Bangla language Facebook comments; compiling a dataset of over 11,000 comments, categorized by polarity (positive, negative, neutral) and various sentiment types, including gender-based hate speech. Furthermore, there are researches conducted on sentiment analysis in the field of online Bangla reviews. For example, \citet{khan2020sentiment} detected depression in Bangla social media using sentiment analysis. They preprocessed a small dataset and employed machine learning classifiers, but faced limitations due to the dataset's size and basic classifiers. 

\citet{akter2021bengali} used machine learning for Bangla e-commerce review sentiment analysis, with KNN achieving 96.25\% accuracy, outperforming other classifiers. This highlighted machine learning's potential in analyzing Bangla e-commerce reviews. Whereas, \citet{banik2018evaluation} introduced a Bangla movie review sentiment analysis system using 800 annotated social media reviews. 
\cite{hasan2023zero} introduced a significant dataset of 33,605 manually annotated Bangla social media posts and examined how different language models perform in zero- and few-shot learning situations. 
Thus, the research of sentiment analysis is continuously growing, and it's helping us better understand sentiment in Bangla online content.

\section{Dataset}

The dataset provided for the shared task \cite{blp2023-overview-task2}, consists of a training set, a development set, and a blind test set. For each set, the texts have been annotated using three labels - 'Positive', 'Neutral', or 'Negative' \cite{islam-etal-2021-sentnob-dataset}. The label distribution for each set is provided in Table \ref{dataset}.

\begin{table} [!h]
\centering
\begin{tabular}{lccc}
\hline
\textbf{Label} & \textbf{Train} & \textbf{Dev} & \textbf{Test} \\
\hline
Positive & 35\% & 35\% & 31\% \\
Neutral & 20\% & 20\% & 19\% \\
Negative & 45\% & 45\% & 50\% \\
\hline
\end{tabular}
\caption{Distribution of instances and labels across training, development, and test sets.}
\label{dataset}
\end{table}

The dataset is imbalanced across the labels, hence it is challenging for the models to learn well.

\section{Experiments}

We conduct a wide range of experiments with several models and data augmentation strategies. Our experiments include statistical models, transformer-based models;  data augmentation strategies like back-translation, multilinguality and also prompting proprietary LLMs. \\

\paragraph{Statistical ML Classifiers} In our experiments, we use statistical machine learning models like Logistic Regression and Support Vector Machine using TF-IDF vectors. We implement both models and some hyperparameter tuning. While SVM performs better with a 0.55 F1 score (Micro) the overall results do not improve much. \\

\begin{figure*} [!h]
  \centering
  \includegraphics[width=\textwidth]{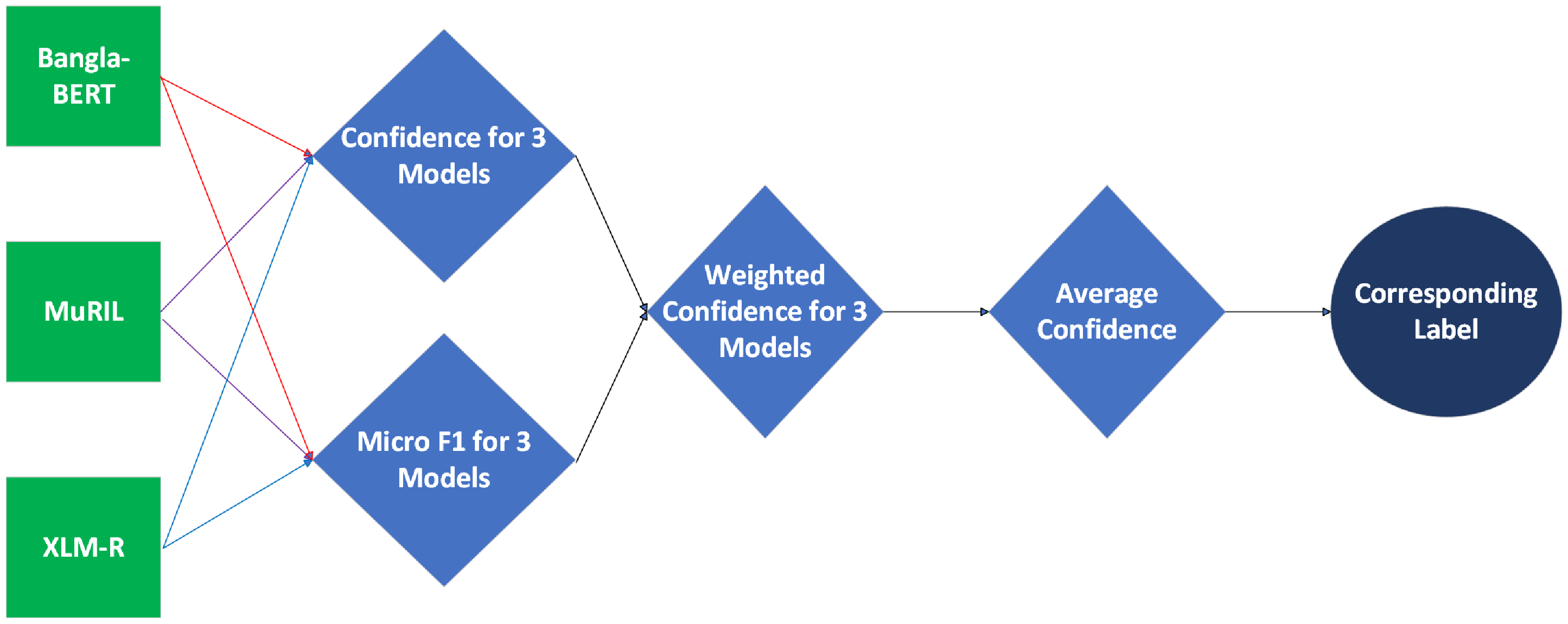}
  \caption{Workflow of the Ensemble Model}
  \label{fig:flowchart}
\end{figure*}

\paragraph{Transformers} We also test several transformer-based models which are pre-trained on Bangla data. Our initial experiments include Bangla-BERT \cite{kowsher2022bangla} which is only pre-trained on bangla corpus. We finetune the model on the train set and evaluate it on the dev set with empirical hyperparameter tuning. We get 0.64 as the best micro F1 using Bangla-BERT.  
We then use multilingual transformer models like multilingual-BERT \cite{devlin2019bert} and xlm-roBERTa \cite{conneau2020unsupervised}, which are pre-trained on 104 and 100 different languages respectively, including Bangla. We also do the same hyperparameter tuning with both models. 
While mBERT gets a 0.60 Micro F1 score, xlm-roBERTa does better with 0.71 on the dev set and 0.70 on the test set. Lastly, we use MuRIL \cite{khanujamuril}, another transformer pre-trained in 17 Indian languages including Bangla. It has a test micro F1 score of 0.67. While experimenting with these models, we observe the losses while fine-tuning to make sure the models do not overfit.

\paragraph{Prompting} Next, we try prompting with gpt-3.5-turbo model \cite{openai2023gpt35turbo} from OpenAI for this classification task. We use the API to prompt the model, while providing a few examples for each label and ask the model to label the dev and test set. The model does not do well with a micro F1 of 0.57 on the dev and 0.51 on the test set.

\paragraph{Transfer Learning on Augmented Data} Finally, we augment the data of the Bangla YouTube Sentiment and Emotion dataset by \citet{hoq2021sentiment}. The dataset has highly positive (2), positive (1), neutral (0), negative (-1) and highly negative (-2) labels. We merge the highly positive and positive labels to Positive, negative and highly negative labels to Negative and keep the neutral label unchanged. This is how we get three labels out of five and merge it with our train data. Following this procedure, we get 0.71 micro F1 score for test dataset.

\begin{table*} [!htb]
  \centering
  \includegraphics[width=\textwidth]{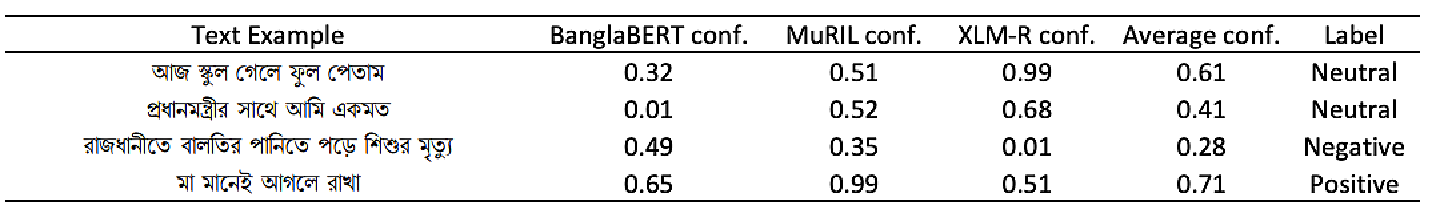}
  \caption{Ensemble with Three Transformer Based Models based on Confidence Score}
  \label{tab:Ensemble}
\end{table*}

\paragraph{Ensemble} \label{data_aug} After finding the results of transformer-based models, we perform an ensemble approach on BanglaBERT, MuRIL, and XLM-R. We then find the weighted average confidence of these three models. For Negative, the confidence interval is fixed 0.0 - 0.33, for Neutral between 0.33 to 0.66 exclusive and for Positive 0.66 - 1.0. The weights are their corresponding test F1 scores found in Table \ref{Results}. With that confidence interval, we predict the test labels. We get a 0.72 micro F1 score by this approach. However this result is not reported to the shared task test phase as we get this result by additional experiments. The detailed label prediction procedure is given in Table \ref{tab:Ensemble} and the workflow of the whole ensemble method is given in Figure \ref{fig:flowchart}. For the first instance, the example is indeed Neutral but BanglaBERT predicts it borderline Negative and XLM-R predicts it Positive. But the power of ensemble approach bring it to the confidence interval of Neutral and thus predicts the label correctly. Similarly, for the second one, a corrected Neutral label is predicted from a Negative, Neutral and borderline Positive confidence. For the last two cases, Negative and Positive labels are determined correctly even with the presence of two Neutral confidence.


\section{Results and Analysis}
At the start of the share task competition, 3 baseline micro F1 scores are provided by the organizers. For random selection the provided baseline is 0.34, for majority selection 0.50, and n-gram 0.55. The results of different models are given in Table \ref{Results}.

\begin{table}[!hb]
\centering
\scalebox{.9}{
\begin{tabular}{lccc}
\hline
\textbf{Models} & \textbf{Dev} & \textbf{Test} \\
\hline
Logistic Regression & 0.47 & 0.45 \\
Support Vector Machine & 0.56 & 0.55 \\
\hline
mBERT & 0.60 & 0.60 \\
BanglaBERT & 0.66 & 0.64 \\
MuRIL & 0.70 & 0.67 \\
XLM-R & 0.71 & 0.70 \\
\hline
GPT 3.5 Turbo & 0.57 & 0.51 \\
\hline
XLM-R (Transfer Learning \\ on Augmented data) & 0.71 & 0.71 \\
\hline
Ensemble & - & \textbf{0.72} \\
\hline
\end{tabular}
}
\caption{Dev and test micro F-1 score for different models and procedures.}
\label{Results}
\end{table}
Amongst the statistical machine learning models, we use logistic regression and support vector machine. For logistic regression, we achieve a micro F1 score of 0.45 and for the support vector machine, the F1 is 0.55.

For transformer-based models, we use mBERT, BanglaBERT, MuRIL and XLM-R where we get the best F1 score of 0.70 by XLM-R.

A few shot learning procedure is used by using GPT3.5 Turbo. We give a few instances of each label as prompt and got 0.51 F1 which is significantly lower than our other attempted approaches except logistic regression. It is because GPT3.5 is still not efficient enough for any downstream classification problem in bangla like this shared task.

Moreover, we augment the data of Bangla YouTube Sentiment and Emotion dataset by \citet{hoq2021sentiment}. The dataset has highly positive, positive labels which we consider as positive and negative, highly negative labels which we consider negative. We keep the neutral label unchanged. This is how we get three labels out of five labels and merge it with our train data. Following this procedure, we finally achieve micro F1 score of 0.71 which we this shared task's leader board.

\begin{figure} [!htb]
\centering
\scalebox{.85}{
\begin{tikzpicture}
\begin{axis}[
    ybar,
    width=1.1\linewidth,
    height=8cm,
    ymin=0, ymax=100,
    symbolic x coords={LR, SVM, mBERT, BanglaBERT, MuRIL, XLM-R,  GPT3.5 Turbo, XLM-R (TL on Augmented Data), Ensemble},
    xtick=data,
    nodes near coords,
    nodes near coords align={vertical},
    x tick label style={rotate=45, anchor=east},
    ]
\addplot[fill=blue!30] coordinates {
    (LR,45)
    (SVM,55)
    (mBERT,60)
    (BanglaBERT,64)
    (MuRIL,67)
    (XLM-R,70)   
    (GPT3.5 Turbo,51)
    (XLM-R (TL on Augmented Data),71)
    (Ensemble,72)
};

\end{axis}
\end{tikzpicture}
}
\caption{Models test Micro-F1 score in percentage.}
\label{fig:Model vs. Performance}
\end{figure}
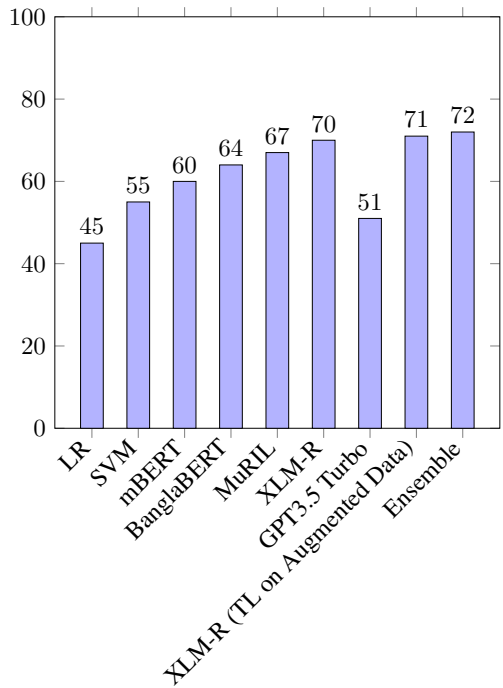

Additionally, we perform ensemble method over the test micro F1 score of BanglaBERT, MuRIL and XLM-R. Instead of doing majority voting on the predicted test label, we find weighted average of confidence interval for the each instances of the test set for the three transformer based models shown in Table \ref{Results}. With that confidence interval, test labels are predicted with 0.72 F1 score which is the best among all our experiments.  A comparison bar chart for different models' performance is shown in  Figure \ref{fig:Model vs. Performance}.

\section{Error Analysis}

\begin{figure*} [!h]
  \centering
  \includegraphics[width=.7\linewidth]{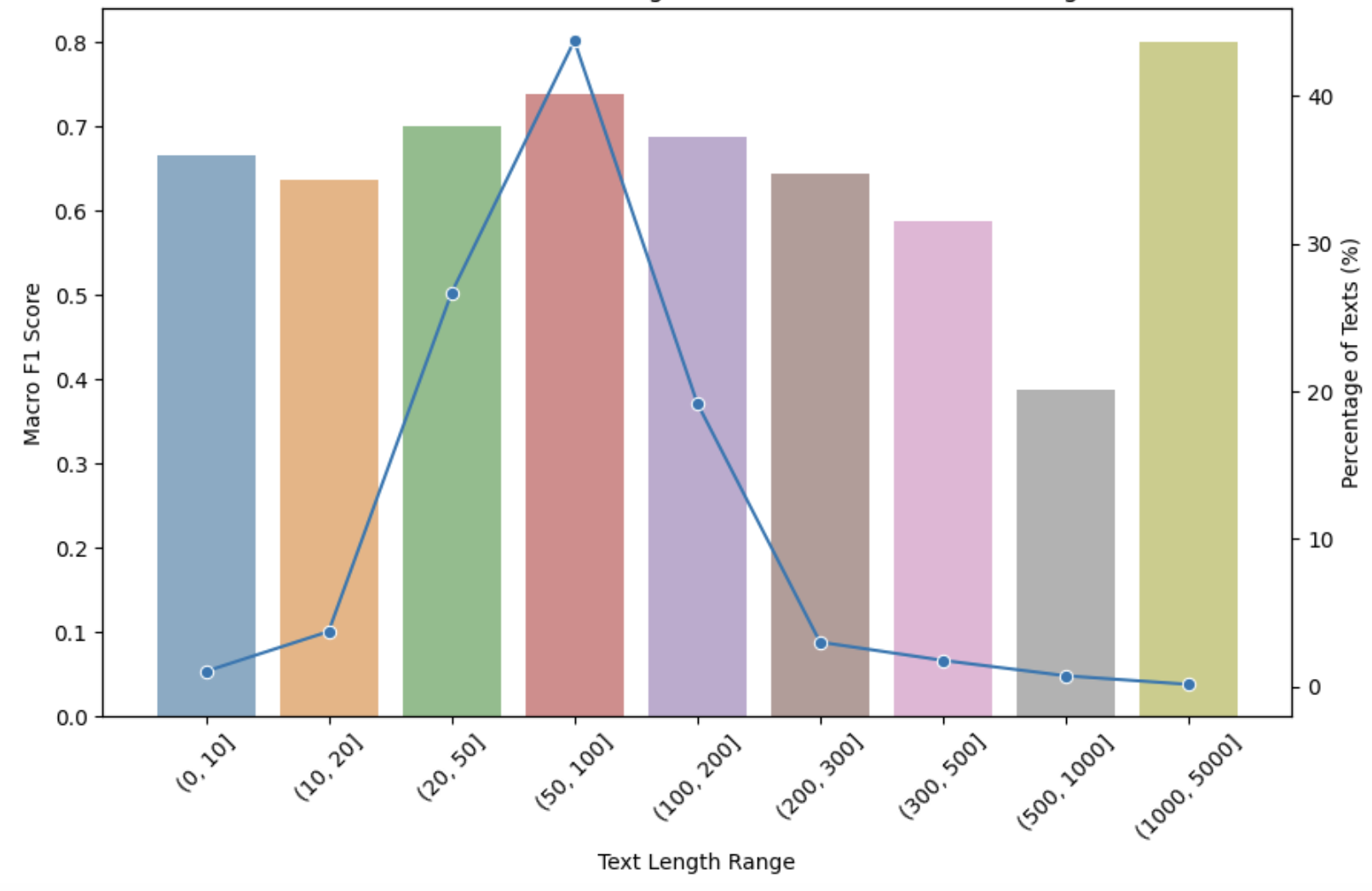}
  \caption{Performance analysis based on text length. }
  \label{fig:Performance Analysis}
\end{figure*}

The classification report provides a comprehensive understanding of our model's performance across the three classes. The overall accuracy of the model is \(0.71\). The 'Positive' class has the highest F1-score of \(0.78\), driven by a precision of \(0.75\) and a recall of \(0.80\). The 'Neutral' class, on the other hand, shows a relatively weaker performance with an F1-score of \(0.42\), a result of its lower precision and recall, \(0.51\) and \(0.37\) respectively. The 'Negative' class offers a competitive performance with an F1-score of \(0.74\), a precision of \(0.72\), and a recall of \(0.76\).

On a macro level, the average values indicate a precision of \(0.66\), recall of \(0.64\), and an F1-score of \(0.65\). When weighted by support, the averages show a slightly better picture with precision at \(0.69\), recall identical to the overall accuracy at \(0.71\), and an F1-score of \(0.70\).

Further dissecting the errors by text length offers more insights. Texts with lengths in the range of \(50\) to \(100\) characters contribute the most to the dataset, constituting \(43.73\%\) of the samples, and have an F1-score of \(0.74\). The second largest group, texts ranging from \(20\) to \(50\) characters, contribute \(26.64\%\) to the dataset with a slightly better F1-score of \(0.70\). It is also worth noting that the performance drastically reduces for texts with lengths between \(500\) and \(1000\) characters, yielding the lowest F1-score of \(0.39\), albeit they only make up \(0.73\%\) of the samples. 

\begin{figure} [!h]
  \centering
  \includegraphics[width=\linewidth]{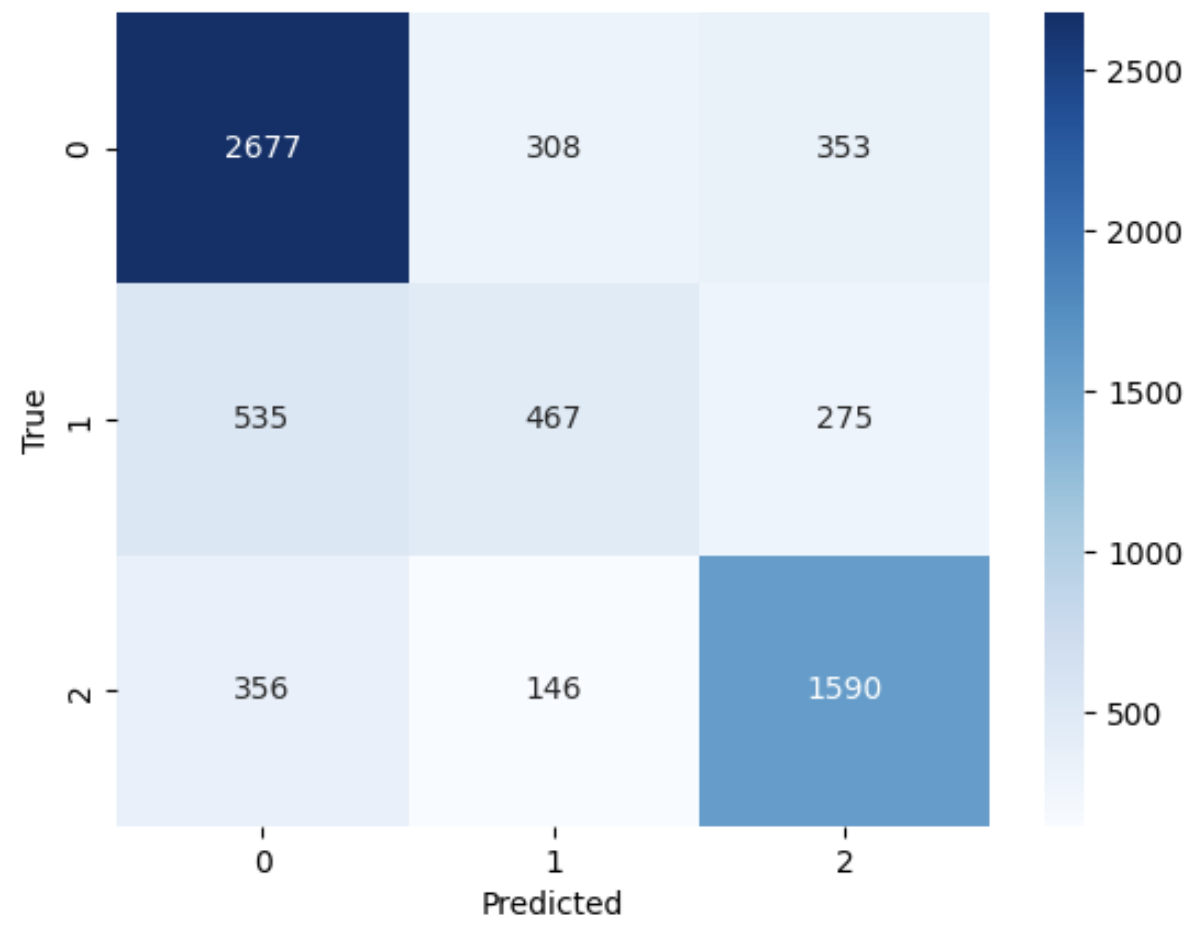}
  \caption{Confusion Matrix}
  \label{fig:confusion matrix}
\end{figure}


\begin{table}[h!]
\centering
\scalebox{.9}{
\begin{tabular}{lcccc}
\hline
Text\_Length & Micro\_F1 & Count & \% \\
\hline
(0, 10] & 0.67 & 69 & 1.03 \\
(10, 20] & 0.64 & 250 & 3.73 \\
(20, 50] & 0.70 & 1787 & 26.64 \\
(50, 100] & 0.74 & 2933 & 43.73 \\
(100, 200] & 0.69 & 1288 & 19.20 \\
(200, 300] & 0.64 & 202 & 3.01 \\
(300, 500] & 0.59 & 119 & 1.77 \\
(500, 1000] & 0.39 & 49 & 0.73 \\
(1000, 5000] & 0.80 & 10 & 0.15 \\
\hline
\end{tabular}
}
\caption{Performance analysis based on text length.}
\end{table}

\section{Conclusion}

In this shared task, we use statistical machine learning models, transformer-based models, a few shot prompting, some customization with transformer-based models with transfer learning, data augmentation, and an ensemble-based approach. The transfer learning and data augmentation procedure is reported as the most successful approach in terms of a micro F1 score of 0.71. But additional experiments by doing an ensemble over three transformer-based models provide a 0.72 F1 score. Overall, this paper can be treated as a holistic experimental outcome for this shared task.

Our transfer learning approach towards solving the problem presented for this shared task shows promising results. However, in most cases, our models keep overfitting. We use dropouts and weight decaying to handle the issue. Even though we perform a lot of hyper-parameter tuning with all the models, it might still be the case that we are not able to find the optimal set of parameters for a few models in our experiments. 

\section*{Acknowledgment}

We would like to thank the shared task organizing for proposing this interesting shared task. We further thank the anonymous workshop reviewers for their valuable feedback. 



\bibliography{anthology,custom}

\begin{thebibliography}{26}
\expandafter\ifx\csname natexlab\endcsname\relax\def\natexlab#1{#1}\fi

\bibitem[{Akter et~al.(2021)Akter, Begum, and Mustafa}]{akter2021bengali}
Mst~Tuhin Akter, Manoara Begum, and Rashed Mustafa. 2021.
\newblock Bengali sentiment analysis of e-commerce product reviews using k-nearest neighbors.
\newblock In \emph{2021 International conference on information and communication technology for sustainable development (ICICT4SD)}, pages 40--44. IEEE.

\bibitem[{Al~Kaiser et~al.(2021)Al~Kaiser, Mandal, Abid, Hossain, Ali, and Naheen}]{al2021social}
Shad Al~Kaiser, Sudipta Mandal, Ashraful~Kalam Abid, Ekhfa Hossain, Ferdous~Bin Ali, and Intisar~Tahmid Naheen. 2021.
\newblock Social media opinion mining based on bangla public post of facebook.
\newblock In \emph{2021 24th International Conference on Computer and Information Technology (ICCIT)}, pages 1--6. IEEE.

\bibitem[{Banik and Rahman(2018)}]{banik2018evaluation}
Nayan Banik and Md~Hasan~Hafizur Rahman. 2018.
\newblock Evaluation of na{\"\i}ve bayes and support vector machines on bangla textual movie reviews.
\newblock In \emph{2018 international conference on Bangla speech and language processing (ICBSLP)}, pages 1--6. IEEE.

\bibitem[{Conneau et~al.(2020)Conneau, Khandelwal, Goyal, Chaudhary, Wenzek, Guzm{\'a}n, Grave, Ott, Zettlemoyer, and Stoyanov}]{conneau2020unsupervised}
Alexis Conneau, Kartikay Khandelwal, Naman Goyal, Vishrav Chaudhary, Guillaume Wenzek, Francisco Guzm{\'a}n, {\'E}douard Grave, Myle Ott, Luke Zettlemoyer, and Veselin Stoyanov. 2020.
\newblock Unsupervised cross-lingual representation learning at scale.
\newblock In \emph{Proceedings of ACL}.

\bibitem[{Devlin et~al.(2019)Devlin, Chang, Lee, and Toutanova}]{devlin2019bert}
Jacob Devlin, Ming-Wei Chang, Kenton Lee, and Kristina Toutanova. 2019.
\newblock {BERT: Pre-training of Deep Bidirectional Transformers for Language Understanding}.
\newblock In \emph{Proceedings of NAACL}.

\bibitem[{Ethnologue(2023)}]{ethnologue2023}
Ethnologue. 2023.
\newblock \href {https://www.ethnologue.com/insights/ethnologue200/} {The most spoken languages worldwide 2023}.

\bibitem[{Haque et~al.(2019)Haque, Manik, and Hashem}]{haque2019opinion}
Fabliha Haque, Md~Motaleb~Hossen Manik, and MMA Hashem. 2019.
\newblock Opinion mining from bangla and phonetic bangla reviews using vectorization methods.
\newblock In \emph{2019 4th International Conference on Electrical Information and Communication Technology (EICT)}, pages 1--6. IEEE.

\bibitem[{Hasan et~al.(2023{\natexlab{a}})Hasan, Alam, Anjum, Das, and Anjum}]{blp2023-overview-task2}
Md.~Arid Hasan, Firoj Alam, Anika Anjum, Shudipta Das, and Afiyat Anjum. 2023{\natexlab{a}}.
\newblock Blp-2023 task 2: Sentiment analysis.
\newblock In \emph{Proceedings of the 1st International Workshop on Bangla Language Processing (BLP-2023)}, Singapore. Association for Computational Linguistics.

\bibitem[{Hasan et~al.(2023{\natexlab{b}})Hasan, Das, Anjum, Alam, Anjum, Sarker, and Noori}]{hasan2023zero}
Md.~Arid Hasan, Shudipta Das, Afiyat Anjum, Firoj Alam, Anika Anjum, Avijit Sarker, and Sheak Rashed~Haider Noori. 2023{\natexlab{b}}.
\newblock \href {http://arxiv.org/abs/2308.10783} {Zero- and few-shot prompting with llms: A comparative study with fine-tuned models for bangla sentiment analysis}.

\bibitem[{Hoq et~al.(2021)Hoq, Haque, and Uddin}]{hoq2021sentiment}
Muntasir Hoq, Promila Haque, and Mohammed~Nazim Uddin. 2021.
\newblock Sentiment analysis of bangla language using deep learning approaches.
\newblock In \emph{International Conference on Computing Science, Communication and Security}, pages 140--151. Springer.

\bibitem[{Iqbal et~al.(2022)Iqbal, Khan, Khan, Iqbal, and Shah}]{iqbal2022sentiment}
Saqib Iqbal, Farhad Khan, Hikmat~Ullah Khan, Tassawar Iqbal, and Jamal~Hussain Shah. 2022.
\newblock Sentiment analysis of social media content in pashto language using deep learning algorithms.
\newblock \emph{Journal of Internet Technology}, 23(7):1669--1677.

\bibitem[{Islam et~al.(2020)Islam, Islam, and Amin}]{islam2020sentiment}
Khondoker~Ittehadul Islam, Md~Saiful Islam, and Md~Ruhul Amin. 2020.
\newblock Sentiment analysis in bengali via transfer learning using multi-lingual bert.
\newblock In \emph{2020 23rd International Conference on Computer and Information Technology (ICCIT)}, pages 1--5. IEEE.

\bibitem[{Islam et~al.(2021)Islam, Kar, Islam, and Amin}]{islam-etal-2021-sentnob-dataset}
Khondoker~Ittehadul Islam, Sudipta Kar, Md~Saiful Islam, and Mohammad~Ruhul Amin. 2021.
\newblock {S}ent{N}o{B}: A dataset for analysing sentiment on noisy {B}angla texts.
\newblock In \emph{Findings of the Association for Computational Linguistics: EMNLP 2021}, pages 3265--3271.

\bibitem[{Kamal et~al.(2016)Kamal, Siddiqi, Afzal, and Rahman}]{kamal2016pashto}
Uzair Kamal, Imran Siddiqi, Hammad Afzal, and Arif~Ur Rahman. 2016.
\newblock Pashto sentiment analysis using lexical features.
\newblock In \emph{Proceedings of the Mediterranean Conference on Pattern Recognition and Artificial Intelligence}, pages 121--124.

\bibitem[{Khan et~al.(2020)Khan, Afroz, Masum, Abujar, and Hossain}]{khan2020sentiment}
Md~Rafidul~Hasan Khan, Umme~Sunzida Afroz, Abu Kaisar~Mohammad Masum, Sheikh Abujar, and Syed~Akhter Hossain. 2020.
\newblock Sentiment analysis from bengali depression dataset using machine learning.
\newblock In \emph{2020 11th international conference on computing, communication and networking technologies (ICCCNT)}, pages 1--5. IEEE.

\bibitem[{Khanuja et~al.(2021)Khanuja, Bansal, Mehtani, Khosla, Dey, Gopalan, Margam, Aggarwal, Nagipogu, Dave et~al.}]{khanujamuril}
Simran Khanuja, Diksha Bansal, Sarvesh Mehtani, Savya Khosla, Atreyee Dey, Balaji Gopalan, Dilip~Kumar Margam, Pooja Aggarwal, Rajiv~Teja Nagipogu, Shachi Dave, et~al. 2021.
\newblock Muril: Multilingual representations for indian languages.

\bibitem[{Kowsher et~al.(2022)Kowsher, Sami, Prottasha, Arefin, Dhar, and Koshiba}]{kowsher2022bangla}
M~Kowsher, Abdullah~As Sami, Nusrat~Jahan Prottasha, Mohammad~Shamsul Arefin, Pranab~Kumar Dhar, and Takeshi Koshiba. 2022.
\newblock Bangla-bert: transformer-based efficient model for transfer learning and language understanding.
\newblock \emph{IEEE Access}, 10:91855--91870.

\bibitem[{Mishev et~al.(2020)Mishev, Gjorgjevikj, Vodenska, Chitkushev, and Trajanov}]{mishev2020evaluation}
Kostadin Mishev, Ana Gjorgjevikj, Irena Vodenska, Lubomir~T Chitkushev, and Dimitar Trajanov. 2020.
\newblock Evaluation of sentiment analysis in finance: from lexicons to transformers.
\newblock \emph{IEEE access}, 8:131662--131682.

\bibitem[{Muhammad and Burney(2023)}]{muhammad2023innovations}
Khalid~Bin Muhammad and SM~Aqil Burney. 2023.
\newblock Innovations in urdu sentiment analysis using machine and deep learning techniques for two-class classification of symmetric datasets.
\newblock \emph{Symmetry}, 15(5):1027.

\bibitem[{Noor et~al.(2019)Noor, Bakhtyar, and Baber}]{noor2019sentiment}
Faiza Noor, Maheen Bakhtyar, and Junaid Baber. 2019.
\newblock Sentiment analysis in e-commerce using svm on roman urdu text.
\newblock In \emph{Emerging Technologies in Computing: Second International Conference, iCETiC 2019, London, UK, August 19--20, 2019, Proceedings 2}, pages 213--222. Springer.

\bibitem[{OpenAI(2023)}]{openai2023gpt35turbo}
OpenAI. 2023.
\newblock \href {https://openai.com/blog/gpt-3-5-turbo-fine-tuning-and-api-updates} {Gpt-3.5 turbo fine-tuning and api updates}.
\newblock Accessed: 2023-08-28.

\bibitem[{Rahman et~al.(2020)Rahman, Pramanik, Sadik, Roy, and Chakraborty}]{rahman2020bangla}
Md~Mahbubur Rahman, Md~Aktaruzzaman Pramanik, Rifat Sadik, Monikrishna Roy, and Partha Chakraborty. 2020.
\newblock Bangla documents classification using transformer based deep learning models.
\newblock In \emph{2020 2nd International Conference on Sustainable Technologies for Industry 4.0 (STI)}, pages 1--5. IEEE.

\bibitem[{Rosenthal et~al.(2017)Rosenthal, Farra, and Nakov}]{rosenthal2017semeval}
Sara Rosenthal, Noura Farra, and Preslav Nakov. 2017.
\newblock Semeval-2017 task 4: Sentiment analysis in twitter.
\newblock In \emph{Proceedings of the 11th International Workshop on Semantic Evaluation (SemEval-2017)}, pages 502--518.

\bibitem[{Saberi and Saad(2017)}]{saberi2017sentiment}
Bilal Saberi and Saidah Saad. 2017.
\newblock Sentiment analysis or opinion mining: A review.
\newblock \emph{Int. J. Adv. Sci. Eng. Inf. Technol}, 7(5):1660--1666.

\bibitem[{Tuhin et~al.(2019)Tuhin, Paul, Nawrine, Akter, and Das}]{tuhin2019automated}
Rashedul~Amin Tuhin, Bechitra~Kumar Paul, Faria Nawrine, Mahbuba Akter, and Amit~Kumar Das. 2019.
\newblock An automated system of sentiment analysis from bangla text using supervised learning techniques.
\newblock In \emph{2019 IEEE 4th International Conference on Computer and Communication Systems (ICCCS)}, pages 360--364. IEEE.

\bibitem[{Yadav and Vishwakarma(2020)}]{yadav2020sentiment}
Ashima Yadav and Dinesh~Kumar Vishwakarma. 2020.
\newblock Sentiment analysis using deep learning architectures: a review.
\newblock \emph{Artificial Intelligence Review}, 53(6):4335--4385.

\end{thebibliography}
\bibliographystyle{acl_natbib}





\end{document}